\documentclass[10pt,twocolumn,letterpaper]{article}

\usepackage{cvpr}              

\usepackage{graphicx}
\usepackage{amsmath}
\usepackage{amssymb}
\usepackage{booktabs}
\usepackage{float}
\usepackage{caption}
\usepackage{colortbl}
\usepackage{xcolor}
\usepackage{float}
\usepackage[export]{adjustbox}
\usepackage[symbol]{footmisc}

\usepackage[pagebackref,breaklinks,colorlinks]{hyperref}

\usepackage[capitalize]{cleveref}
\crefname{section}{Sec.}{Secs.}
\Crefname{section}{Section}{Sections}
\Crefname{table}{Table}{Tables}
\crefname{table}{Tab.}{Tabs.}

\newcommand{\cL}{\mathcal{L}}

\makeatletter
\DeclareRobustCommand\onedot{\futurelet\@let@token\@onedot}
\def\@onedot{\ifx\@let@token.\else.\null\fi\xspace}

\makeatother

\newcommand{\PAR}[1]{\vspace{0.1cm}\noindent{\bf #1} }

\begin{document}
	
\title{SCL-VI: Self-supervised Context Learning for Visual Inspection of Industrial Defects}

\vspace{-2cm}

\author{Peng Wang$^{1,2}$\footnotemark[1]\qquad Haiming Yao$^{3}$\qquad Wenyong Yu$^{4}$ \\$^{1}$Westlake University\qquad $^{2}$Zhejiang University\qquad $^{3}$Tsinghua University\qquad $^{4}$HUST}


%
\maketitle
\footnotetext[1]{work was done at HUST.}
%

\begin{abstract}
The unsupervised visual inspection of defects in industrial products poses a significant challenge due to substantial variations in product surfaces. Current unsupervised models struggle to strike a balance between detecting texture and object defects, lacking the capacity to discern latent representations and intricate features. In this paper, we present a novel self-supervised learning algorithm designed to derive an optimal encoder by tackling the renowned jigsaw puzzle. Our approach involves dividing the target image into nine patches, tasking the encoder with predicting the relative position relationships between any two patches to extract rich semantics. Subsequently, we introduce an affinity-augmentation method to accentuate differences between normal and abnormal latent representations. Leveraging the classic support vector data description algorithm yields final detection results. Experimental outcomes demonstrate that our proposed method achieves outstanding detection and segmentation performance on the widely used MVTec AD dataset, with rates of 95.8 \% and 96.8 \%, respectively, establishing a state-of-the-art benchmark for both texture and object defects. Comprehensive experimentation underscores the effectiveness of our approach in diverse industrial applications. Code is available at \href{https://github.com/wangpeng000/VisualInspection}{https://github.com/wangpeng000/VisualInspection}.
\end{abstract}


\section{Introduction}
\label{sec:intro}
Visual inspection, owing to its remarkable attributes such as non-contact functionality, speed, and adaptability, has found extensive application in Automatic Optical Inspection (AOI) for detecting defects in various industrial products. This includes applications in diverse fields such as blades \cite{bib1}, thin-film transistor liquid crystal displays \cite{bib2}, and robot parts \cite{bib3}.

Nonetheless, the visual inspection of surface defects in products continues to pose challenges due to factors such as diverse sizes, varying brightness, low contrast, complex features, and the insufficient availability of defect samples across different products. Over the past few decades, researchers have introduced numerous methods for surface defect inspection to address these challenges. These methods can be broadly categorized into two fundamental groups based on their feature extraction strategy: traditional methods and deep learning methods.

Traditional methods typically employ manually designed features for texture defect inspection. For instance, Tolba et al. \cite{bib4} introduced the gray-level cooccurrence method, leveraging statistical analysis for product surface feature assessment. Aiger and Talbot \cite{bib5} utilized pure spectral phase conversions for defect detection. However, these traditional approaches heavily depend on handcrafted features. Essentially, as the number of exceptions and defect classes grows, the reliance on prior experience renders these traditional methods inadequate to meet the automation requirements of a factory.

In recent times, deep learning methods have gained widespread adoption in the realm of visual inspection owing to their capacity to efficiently extract highly complex features. These methods can be categorized into supervised and unsupervised approaches based on the presence of labels in the training data. Supervised learning, while effective, demands substantial time and labor for data labeling, presenting challenges in acquiring a sufficient number of defective samples in real industrial production settings. On the other hand, unsupervised learning methods for defect inspection only require defect-free samples without labels and can be broadly classified into two major types: reconstruction-based and objective function optimization-based strategies \cite{bib6}.

Li \cite{bib7} and Yang et al. \cite{bib8} employed reconstruction-based methods for inspecting defects on textural surfaces, utilizing autoencoders (AE) and generative adversarial networks (GAN). These methods aim to indirectly target abnormal defects by reconstructing the background and subtracting the original images from the reconstructed ones. However, a drawback is the potential inclusion of false information in the generated images. An alternative approach is to directly focus on anomalies \cite{bib9}. Methods like support vector data description (SVDD) \cite{bib10} fall into this category, where normal data is clustered in a compact latent space, minimizing the objective function loss to ensure separation from abnormal data.

For instance, Ruff \cite{bib6} integrated a deep neural network into SVDD for detecting abnormal images. However, a limitation is the inability to distinguish similar features from different areas of products, as they are all encompassed in a single latent feature space. Addressing this, Yi \cite{bib11} proposed a patch-level SVDD method that generates multiple SVDD latent representation spaces by inputting patch-level images into a deep network optimized through self-supervised learning \cite{bib12}. Nevertheless, the self-supervised method in \cite{bib11} has limitations in central areas, leading to the loss of certain semantics.

This paper introduces a novel context self-supervised method designed to efficiently detect various defects with training using only a small number of normal samples. Drawing inspiration from the advantages of patch-level Support Vector Data Description (SVDD) \cite{bib11}, our method can encapsulate complex latent representations within different hyperspheres rather than being confined to a singular one. The key {\bf contributions} of our work include:
\begin{itemize}
	\item we peopose a novel and effective self-supervised learning method to optimize the parameters of deep networks;
	\item developing a comprehensive affinity-augmentation mechanism with broad applicability, aimed at enhancing the distinction between normal and abnormal features to facilitate defect identification;
	\item We deliver better results on MvTec and other real dataset.
\end{itemize}

\section{Related Work}
\label{sec:related}
In this section, the related works of self-supervised learning and SVDDs are introduced and discussed.

\PAR{Self-supervised semantic learning.} As a recent evolution in the realm of unsupervised learning, self-supervised learning has demonstrated remarkable success in diverse applications, including image restoration \cite{bib13}, image segmentation \cite{bib14}, image detection \cite{bib12}, \cite{bib15}, \cite{bib16}, scene description \cite{bib17}, and target tracking \cite{bib18}. Self-supervised learning primarily leverages pretext tasks to derive informative labels and semantics from the data, thereby enhancing performance in downstream tasks. Despite their prevalence in defect detection methods, Autoencoders (AEs) and Generative Adversarial Networks (GANs), such as CAE \cite{bib19}, DAE \cite{bib20}, AFEAN \cite{bib8}, and F-AnoGAN \cite{bib21}, tend to overlook crucial semantic relationships between pixels. There is a need for more advanced models to capture richer semantics in defect detection.

Semantic learning constraints in self-supervised learning can be broadly categorized into temporal-based \cite{bib17}, \cite{bib18}, and context-based \cite{bib12}, \cite{bib13}, \cite{bib14}, \cite{bib15}, \cite{bib16} approaches. Sermanet \cite{bib17} and Wang \cite{bib18} employed self-supervised constraint training based on frame similarity within a video sequence. These temporal-based self-supervised methods focus on learning the semantics of objects at different time points, primarily finding applications in video processing and language systems. However, our work is particularly relevant to the consideration of semantics across different parts of an object.

Previous studies have successfully demonstrated that context-based learning is effective in extracting rich semantics from images \cite{bib13}. For instance, Pathak et al. \cite{bib13} employed a context encoder to predict the removed patch, contributing to representation learning. However, this approach did not fully capitalize on representation learning. Noroozi proposed a model to acquire feature representations of an image through solving a jigsaw puzzle \cite{bib14}. While emphasizing differences, this method overlooked similarities among patches. Gidaris \cite{bib15} and Lee \cite{bib16} introduced rotation and RGB channel transformations, respectively, to enable the model to comprehend object properties, such as position and color. Additionally, Doersch \cite{bib12} demonstrated the significant enhancements in image classification and recognition achievable through context learning, particularly by predicting the relative positions of the center and the surrounding eight blocks. However, due to the central area's limitations, this method overlooked certain semantics embedded in the patches.

The self-supervised context learning method presented in this paper aligns with the overarching objective of Doersch's approach \cite{bib12}, aiming to capture high-level semantics. Notably, our method introduces a random selection of two patches from the nine available, effectively mitigating the limitations associated with the central area. This deliberate patch selection strategy contributes to a balanced consideration of both similarity and difference by focusing on the relative positions of the chosen patches.

\PAR{SVDDs.} One-class classification \cite{bib10}, \cite{bib22} represents an unsupervised approach for segregating data into normal and abnormal categories. Tax et al. \cite{bib10} enhanced One-Class Support Vector Machines (OC-SVM) by introducing Support Vector Data Description (SVDD). SVDD aims to identify the center and radius of the smallest hypersphere containing normal data, utilizing this information to separate data using a hypersphere rather than a hyperplane. Notably, when employing a Gaussian kernel, OC-SVM and SVDD have been demonstrated to be equivalent \cite{bib23}. The use of manually designed kernels has been identified as a factor contributing to poor computational scaling performance for OC-SVM and SVDD \cite{bib11}, \cite{bib24}.
\begin{figure*}
	\centering
	\includegraphics[width=0.9\linewidth]{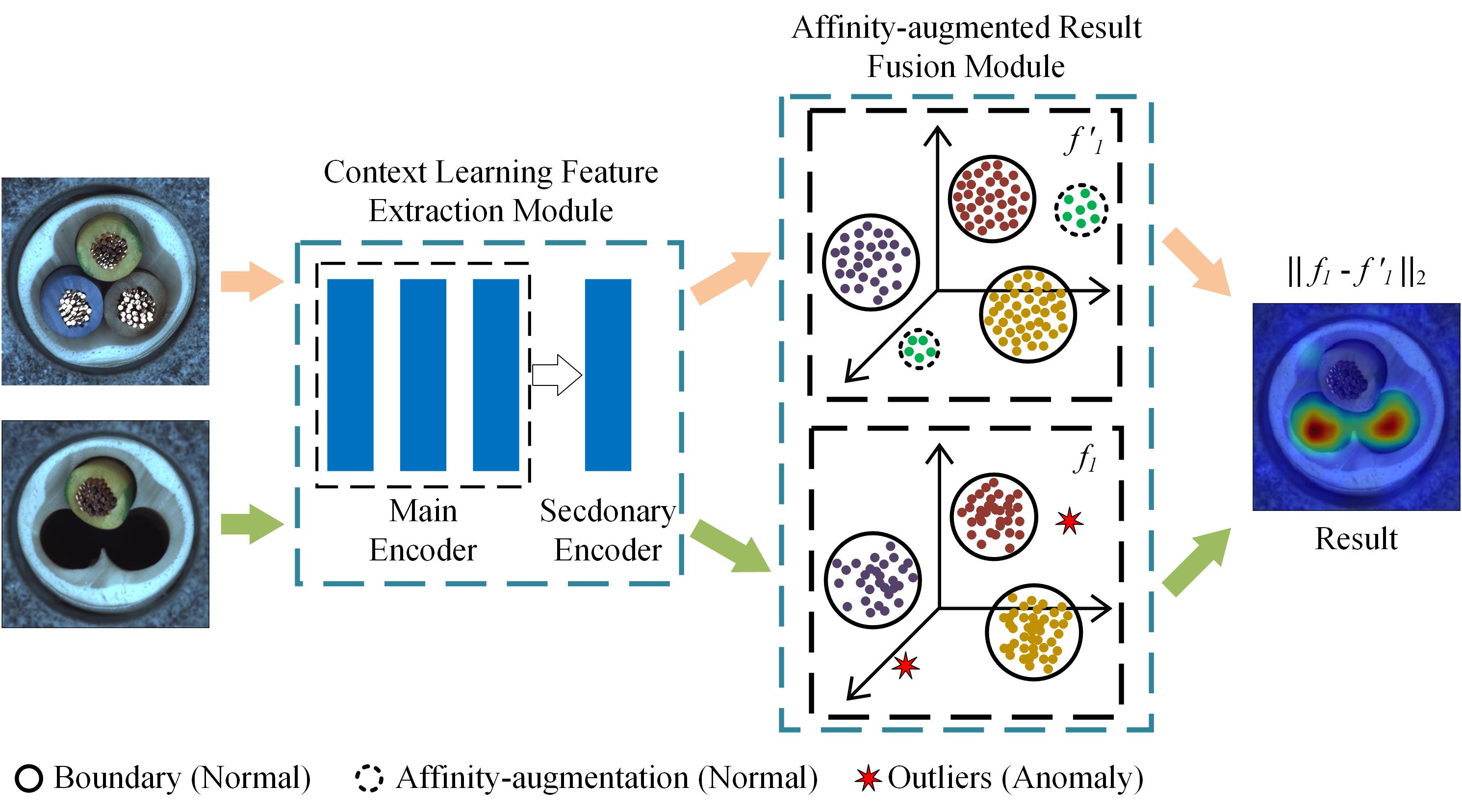}
	\caption{{\bf{Overview.}} we incorporate the fundamental concept of patch-level Support Vector Data Description (SVDD) from \cite{bib11}. The context learning feature extraction module comprises two encoders constructed through multiple convolutional layers. The initial layers form the primary encoder, utilized for both large and small-sized inputs, while the subsequent layers constitute the secondary encoder, exclusively employed for large-sized inputs. These two encoders leverage our novel context learning method to acquire rich semantics, endowing them with robust semantic coding capabilities. They extract diverse features from patches of different sizes, enhancing their capacity to detect defects of various sizes. Following this, we apply the classic patch-level SVDD method from \cite{bib11}, encapsulating normal features into hyperspheres and positioning abnormal features outside these hyperspheres. Each defect undergoes inspection by comparing anomaly features with affinity-augmented normal features stored in memory, incorporating calculated weights. Finally, anomaly maps of varying sizes are fused to yield precise defect detection results.}
	\label{fig1}
\end{figure*}

To overcome these challenges, Ruff et al. \cite{bib6} introduced the deep SVDD method, leveraging a neural network to autonomously train and determine the center and radius of the hypersphere. This approach eliminates the necessity for manually selecting an appropriate kernel. Notably, a closely related work to ours is Patch-SVDD \cite{bib11}, which creatively integrated a self-supervised method with SVDD, utilizing patches instead of the entire image to establish multiple hyperspheres encompassing features from various regions of the image. In our research, we build upon this patch-level SVDD framework, adapting its self-supervised aspect, and introducing an affinity-augmentation mechanism to further enhance the inspection results for images.

\section{Method}\label{sec:method}
We introduce a self-supervised method designed for the detection of diverse industrial product defects within a unified feature domain. Our proposed approach establishes memory items during the training phase to learn and store normal features using only a small number of defect-free samples. The overall structure of our method comprises three modules, as depicted in Figure~\ref{fig1}: a novel context learning feature extraction module, a patchwise feature discrimination module (refer to \cite{bib11}), and an affinity-augmented result fusion module.

\subsection{Context Learning Feature Extraction Module}
\begin{figure*}
	\centering
	\includegraphics[width=0.9\linewidth]{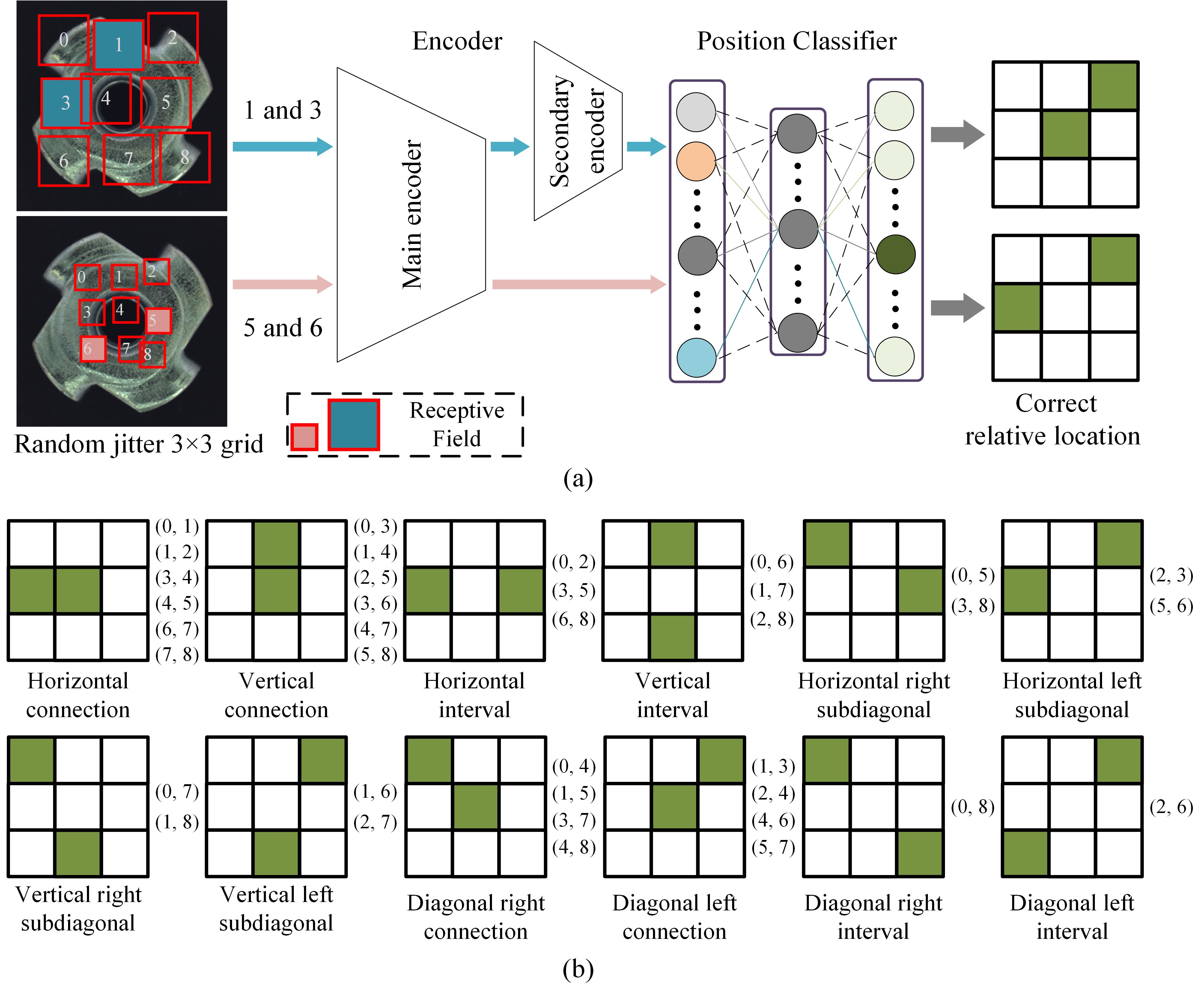}
	\caption{{\bf{Context Learning Module.}} A novel context prediction self-supervised learning method. (a) Multiscale context prediction self-supervised learning. (b) Twelve context prediction positions. The two dark grid squares in each 3×3 grid represent the relative positions. The numbers in brackets represent the absolute positions of the two selected patches.}
	\label{fig2}
\end{figure*}

The feasibility of learning semantics through predicting the relative positions of patches has been well-established \cite{bib12}. In an effort to enhance the encoder's capability to capture semantics and improve feature discrimination, we introduce a novel self-supervised learning method, as illustrated in Figure~\ref{fig2}.
In our pretext task, two patches, denoted as $p_1$ and $p_2$, are randomly selected within the 3$\times$3 kernel space in each iteration. The correct relative distribution of these two patches is determined through the training of the encoder and classifier, as illustrated in Figure~\ref{fig2}(a). There exist $C_9^2=36$ cases of absolute position relationships between $p_1$ and $p_2$. However, it's noteworthy that one relative position relationship encompasses multiple absolute position relationships. For instance, the relative position relationship is the same for pairs like 1 and 3, 2 and 4, 4 and 6, and 5 and 7. These four absolute positions belong to the same relative position of the diagonal left connection and share similar semantics. Consequently, only twelve distinct relative position relationships are considered among the 36 absolute relationships, as depicted in Figure~\ref{fig2}(b). The relative positions are encoded as $y\in\{0,1,\ldots,10,11\}$, and the encoder $E$ and classifier $C$ predict the relative positions of $p_1$ and $p_2$ when input into the model. The loss for self-supervised learning is defined as follows:
\begin{equation}
	\cL_{SSL}={CrossEntropy}\left(y, C\left(E\left(p_{1}\right), E\left(p_{2}\right)\right)\right)\label{eq1}
\end{equation}

Our algorithm deviates from those presented in \cite{bib11} and \cite{bib12}. In comparison with Doersch's method \cite{bib12}, our proposed pretext task captures more semantics and is not confined to determining the relative positions between a specific block and its eight surrounding blocks. During our experiments, to mitigate the classifier's reliance on shortcuts, such as patch boundary information and color details, we introduce random dithering of 0-16 pixels for each patch and adjust the RGB value by randomly increasing or decreasing the gray value. This ensures that the classification is rooted in semantics rather than shortcuts. As a result, the encoder demonstrates improved feature discrimination ability following training with self-supervised learning.

\subsection{Patchwise Feature Discrimination Module}
Combing deep SVDD \cite{bib6} with a patch-level idea, \cite{bib11} trained a deep patch SVDD network, which outputs multiple smaller hyperspheres with patchwise data as the input. Each hypersphere contains the features of a specific region. In our approach, we regard patch-SVDD as a basic tool and our baseline.

In the training phase, the function of the encoder in patch SVDD is to wrap the features of the patches in the same position into one hypersphere as much as possible, that is, to minimize the feature distance between the patches. \cite{bib11} defines the distance loss as follows :
\begin{equation}
	\cL_{pSVDD}=\sum_{i}^{N}\left\|E\left(p_{i}\right)-E\left(p_{i}^{'}\right)\right\|_{2}, i=1,2, \ldots, N\label{eq2}
\end{equation}

where $p_i$ and $p_i^{'}$ are two extremely similar patches and $p_i^{'}$ is randomly offset by several pixels on the basis of $p_i$ during training.

The fusion loss of patch SVDD and self-supervised learning can be expressed as follows \cite{bib11}:
\begin{equation}
	\cL=\cL_{SSL}+\alpha \times \cL_{pSVDD}\label{eq3}
\end{equation}

where $\alpha$ in the formula adjusts the contributions of the different semantics, and the main loss $\cL_{SSL}$ is our new self-supervised learning loss.

\subsection{Affinity-Augmented Fusion Module}
In the baseline method (patch SVDD) \cite{bib11}, using only the nearest feature as the normal feature can create a perception that the discovered image feature is not the most suitable due to variations in normal data features, leading to the issue of contingency. Conversely, taking the average of multiple normal features might result in the averaged feature being less similar to the nearest normal feature due to substantial feature differences. The process of obtaining normal features significantly influences the final results.

To address both the problem of contingency and differences in feature comparison, normal features are collectively stored in a structure referred to as a memory item. Assuming the Euclidean distance between the $N$ normal compositional patterns $(p_1, p_2, \ldots, p_N)$ retrieved from the memory module and the latent representation $(p)$ being tested is $d_1, d_2, \ldots, d_N$ $(d_i=\|p_i-p\|_2)$, the affinity $\beta_i$ is defined as follows:
\begin{equation}
	\beta_{i}=\frac{e^{\gamma_{i}}}{e^{\gamma_{1}}+e^{\gamma_{2}}+\cdots+e^{\gamma_{N}}}, \quad i=1,2, \ldots, N\label{eq4}
\end{equation}
where $\gamma_{i}=\min \left\{\frac{d_{1}+d_{2}+\cdots+d_{N}}{d_{i}}, \lambda\right\}$. The threshold $\lambda$ prevents $d_i$ from being infinitely small enough such that $e^{\gamma_{i}}$ leads to computation overflow. Therefore, $\lambda$ as a value in the interval [15, 30] is recommended. In this paper, $\lambda$ is set to 20.

$N$ multiple normal compositional patterns(four in Figure~\ref{fig3}) are fused to obtain a fusion pattern $p_{fused}$ :
\begin{equation}
	p_{fused}=\beta_{1} p_{1}+\beta_{2} p_{2}+\cdots+\beta_{N} p_{N}\label{eq5}
\end{equation}

After the training phase, the encoder has learned the ability to map normal samples into several hyperspheres. Each hypersphere can represent the basic microstructure of the normal sample. In the test phase, the abnormal score $score(p)$ is obtained by calculating the distance between the latest representation of the test pattern and the fusion pattern. which is an improvement over the baseline, as follows:
\begin{equation}
	{score}(p)=\left\|E(p)-E\left(p_{fused}\right)\right\|_{2}\label{eq6}
\end{equation}

A pixel belongs to multiple patches because the sliding stride of the patch is not equal to the width of the patch. Therefore, the anomaly score of each pixel $pixel_{\delta}$ depends on the average value of the abnormality score of the $K$ patches to which it belongs:
\begin{equation}
	{score}\left({pixel}_{\delta}\right)=\frac{1}{K} \sum_{j}^{K}{score}\left(p_{j}\right), j=1,2, \ldots, K
\end{equation}\label{eq7}
where $p_j$ represents the $K$ patches to which the pixel belongs. Pixels with relatively high scores are considered defects.

\section{Experiments}
\label{sec:experiment}
\subsection{Experimental details}
To assess the effectiveness of the proposed method, we conduct various sets of experiments in this section. The key hyperparameter $\alpha$ has been explored in \cite{bib11}; hence, this paper does not delve into discussing $\alpha$. First, we illustrate the boosting self-supervised learning paradigm that parses context information. Second, we examine the impact of the number of memory items associated with affinity. Following this, we compare the overall performance of our method with several state-of-the-art models. Lastly, we deploy the model in a real industrial environment for practical use. 

\PAR{Benchmark datasets.} In our experiments, we leverage various types of samples from the MVTec AD dataset. MVTec AD is a comprehensive dataset comprising 15 different industrial products found in the real world, featuring manually generated defects that mimic those encountered in real-world industrial inspection scenarios. This dataset is widely employed for performance evaluation. The training set encompasses 3629 defect-free images, incorporating 5 textural materials (carpet, grid, leather, tile, wood), and 10 object products (bottle, cable, capsule, hazelnut, metal-nut, pill, screw, toothbrush, transistor, zipper). The testing sets include 1725 images, consisting of both defective and defect-free samples, featuring 73 distinct types of defects.

\PAR{Basrline.} The performance of the proposed method on pretext tasks is quantitatively compared with that of famous state-of-the-art models (Jigsaw1 (Baseline) \cite{bib11}, \cite{bib12}, Jigsaw2 \cite{bib14}, Rotation \cite{bib15}, RGB \cite{bib16}, and Rotation+RGB (RR) \cite{bib16}). 

\PAR{Implementation details.} To quantitatively compare the performance among the different methods, we adopt a threshold-independent evaluation indicator, the area under the receiver operating characteristic curve (AUROC), which is widely used to evaluate the performance of binary classifiers. All of the experiments are conducted with PyTorch. The main architecture of the encoder in the experiment is shown in Table~\ref{tabA1} and Table~\ref{tabA2}, and GroupNorm is used after each convolutional layer for convergence stability. To enable a comparison with the baseline (patch SVDD \cite{bib11}), all input images are scaled to 256×256 pixels, and the patch sizes are set to 64×64 and 32×32.
\subsection{Ablation study}
\PAR{Influence of context self-supervised learning.} An ablation experiment on the novel self-supervised learning paradigm proposed in this paper is described in this section. We regard patch SVDD as the baseline, change its self-supervised algorithm and ensure that other conditions are completely consistent, to compare the performance with that of various self-supervised algorithms. Since the jigsaw self-supervised method in \cite{bib12} is used in patch SVDD, among the results of the comparison, \cite{bib12} represents the baseline. The experimental results are shown in Table~\ref{table_ssl} and the detailed data are shown in Table~\ref{tabA3} and Table~\ref{tabA4} in the supplementary materials (Sec.~\ref{sec:sup}).
\begin{table}
	\setlength\tabcolsep{5pt}
	\centering
	\resizebox{\linewidth}{!}{
		\begin{tabular}{c|cccccc|c}
			\specialrule{0.1em}{1pt}{1pt}
			& \footnotesize NoSSL	& \footnotesize Jigsaw1	& \footnotesize Jigsaw2
			& \footnotesize Rotation	& \footnotesize RGB & \footnotesize RR
			& \footnotesize \bf{Ours}\\
			\hline
			\scriptsize {Detection}
			& \footnotesize73.9 & \footnotesize92.1 & \footnotesize88.0
			& \footnotesize89.8 & \footnotesize88.3 & \footnotesize89.6
			& \footnotesize \bf{95.3}\\
			\scriptsize {Segmentation}
			& \footnotesize88.1 & \footnotesize95.7 & \footnotesize94.5
			& \footnotesize92.1 & \footnotesize94.1 & \footnotesize93.6
			& \footnotesize \bf{96.6}\\
			\specialrule{0.1em}{1pt}{1pt}
		\end{tabular}
	}
	\vspace{-0.8em}
	\caption{{\bf{Contrast of self-supervised learning.}} The self-supervised learning strategy provides a novel way to boost detection and the corresponding segmentation performance in the metric of AUCROC (\%).}
	\label{table_ssl}
\end{table}
\begin{figure}[t]%
	\centering
	\includegraphics[width=0.45\textwidth]{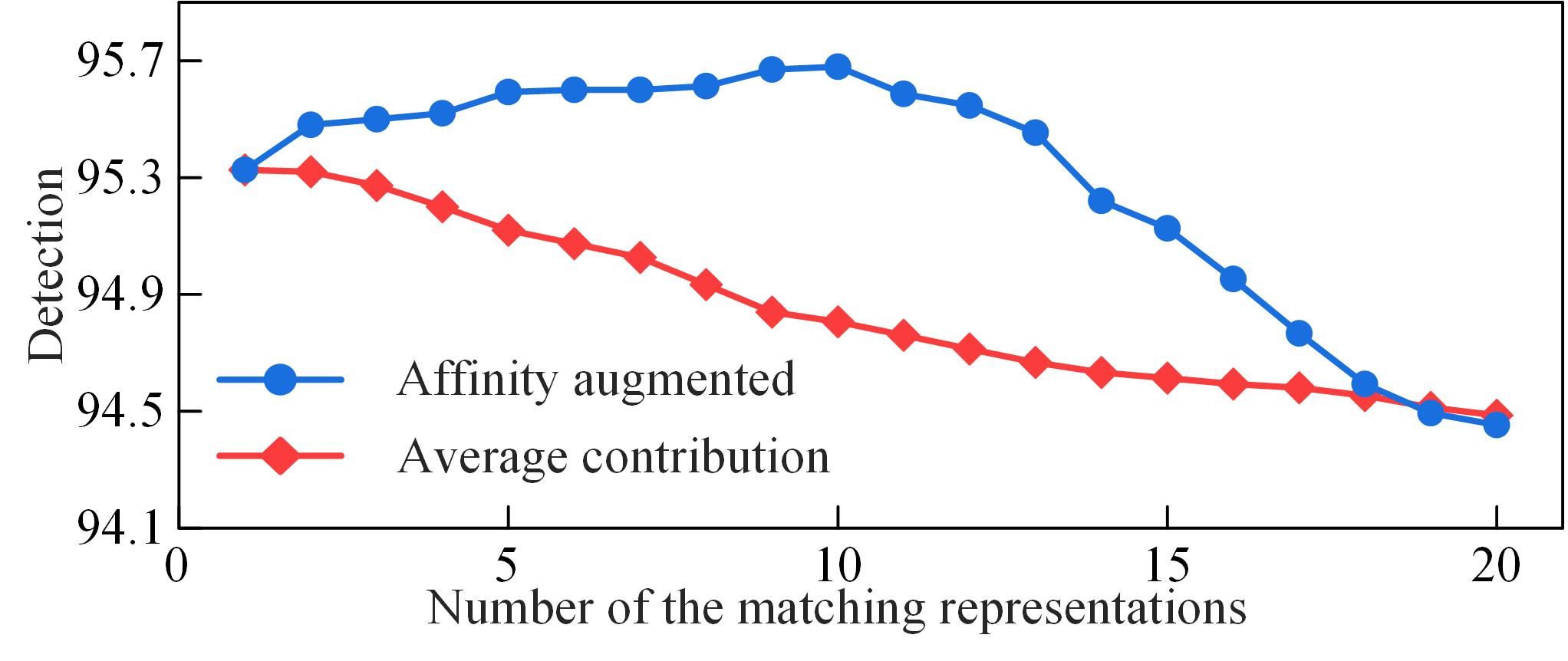}
	\caption{Influence of the number of matching items ($\eta$) on detection}\label{fig6}
\end{figure}
For the outstanding method \cite{bib11}, \cite{bib12}, correct position identification of the scrambled patches is employed. For homogeneous textures, different adjacent relative positions may have similar characteristics; in other words, context information is easy to omit due to the shortcut that connectivity between textures is easier to extract. For object products, the receptive field is limited by the local patch locations; thus, only partial semantic information can be learned. In contrast, the proposed pretext task is innovative not only in randomly selecting two patches in the 3$\times$3 kernel space, eliminating the "shortcut" problem but also in dividing the relative positions into 12 representative distributions. It exhibits relatively good performance due to the robust representations.

\PAR{Influence of the number of affinity.} The affinity mechanism is designed to identify anomaly representations in the latent space. The parameter $\eta$ in formula~\ref{eq4}, representing how many memory items are used to obtain $p_{fused}$ , is a crucial factor influencing the inspection results. We aim to explore the sensitivity of the parameter $\eta$, as depicted in Figure~\ref{fig6}. The detection and segmentation performance of the proposed method on the MVTec AD datasets are plotted in terms of the Area Under the Receiver Operating Characteristic curve (AUROC), varying the value of $\eta$. The trends indicate that as $\eta$ increases, the average contribution of the mechanism decreases. In contrast, the affinity mechanism first increases and then decreases, reaching its maximum AUROC value when $\eta=10$. This is because, during the testing phase, the anomaly score of each representation is calculated using normal sample representations with more affinity. Adopting a single representation strategy yields poorer results due to contingency. The synthesis of results from the affinity mechanism proves beneficial in extracting more semantic information from normal samples. However, introducing interference from irrelevant items occurs with a $\eta$ value that is too large. In this paper, $\eta$ is set to 10 to strike a balance, considering both the affinity mechanism's effectiveness and the potential interference introduced by irrelevant items.
\subsection{Overall performance comparison}
To comprehensively evaluate the effectiveness of the proposed method, its detection and segmentation performance is compared with several state-of-the-art methods, including autoencoders (AE-SSIM \cite{bib25}, MEM-AE \cite{bib26}, and Trust-MAE \cite{bib27}), generative adversarial networks (F-Aon-GAN \cite{bib21}, AnoGAN \cite{bib28}), the variational method VAE-GRAD \cite{bib29}, and other superior unsupervised algorithms (Patch SVDD \cite{bib11}, US \cite{bib30}, RIAD \cite{bib31}). The hyperparameter settings are determined based on ablation experimental results; specifically, $\alpha$ is set to 0.0001, and $\eta=10$ in subsequent experiments.

\begin{figure*}[t]
	\centering
	\includegraphics[width=0.77\textwidth]{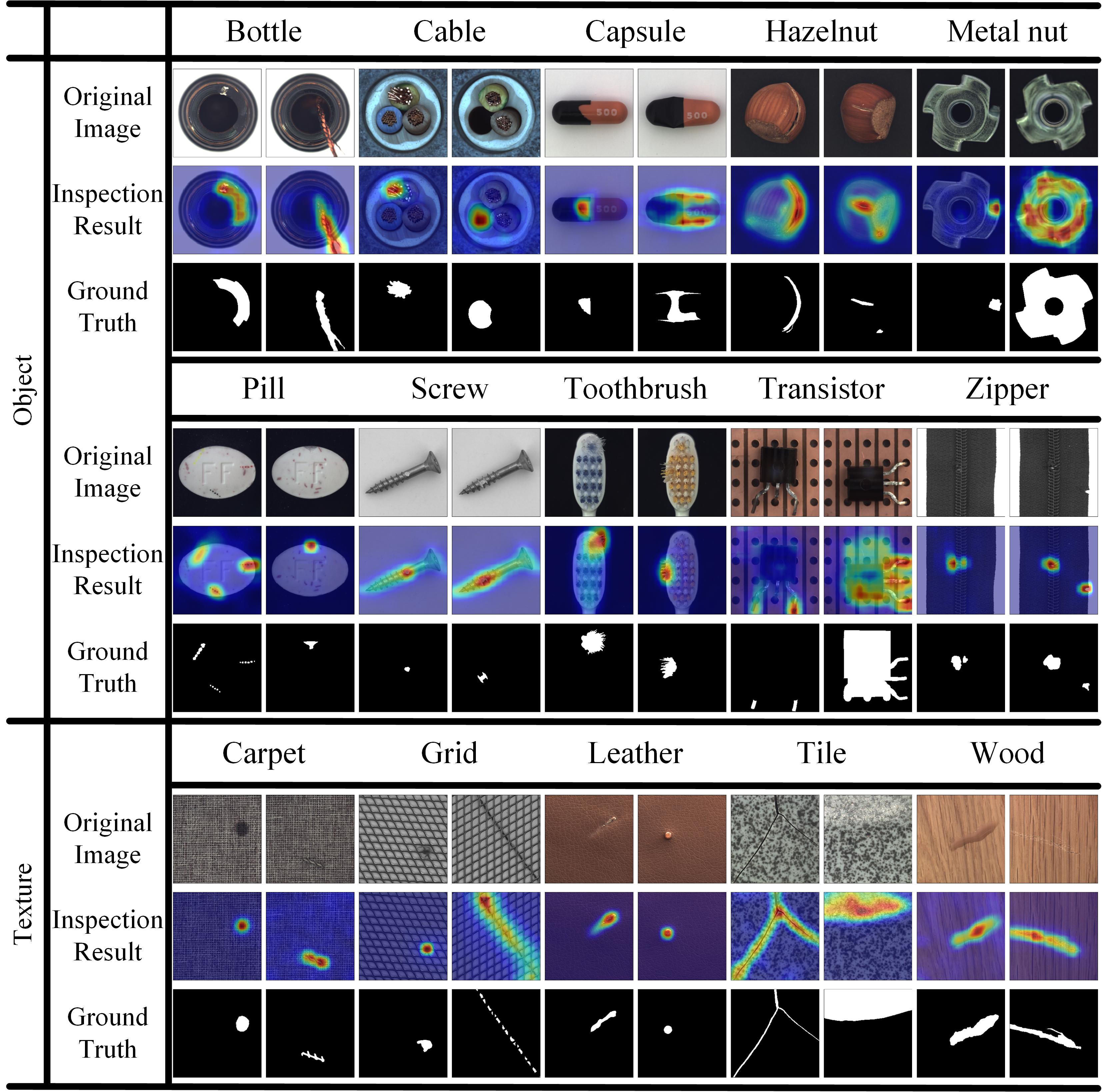}
	\caption{Inspection results of our proposed method on MVTec AD.}\label{fig8}
\end{figure*}

The detection and segmentation results of these methods on the MVTec AD dataset are presented in Table~\ref{tabA5} and Table~\ref{tabA6}. As observed in Table~\ref{tabA5}, the proposed method attains the best AUROC performance in detection. Compared to the best existing method, the proposed method improves the detection performance by a margin of 3.74\%. The segmentation results in Table~\ref{tabA6} demonstrate margins of 1.06\% and 2.61\% when compared with the best and second-best methods. The proposed method exhibits excellent performance across both textured surfaces and object products, showcasing its superiority, stability, and potential to serve as a unified model for defect inspection in industrial applications.
\begin{table*}\large
	\parbox{\textwidth}{
		\resizebox{0.97\linewidth}{!}{
	\begin{tabular}{cccccccccc}
		\toprule%
		Category&{\bf{Ours}} & F-AnoGAN & MEMAE & AESSIM & TrustMAE &VAEGRAD &PatchSVDD &US &RIAD\\
		\midrule
		Carpet & 96.4 & 56.6 & 74.6 & 87.0 & \textbf{97.4} & 74.0 & 92.9 & 91.6 & 84.2  \\
		Grid & 97.0 & 59.6 & 98.9 & 94.0 & 99.1 & 96.0 & 94.6 & 81.0 & \textbf{99.6}  \\
		Leather & 99.2 & 62.5 & 95.5 & 78.0 & 95.1 & 93.0 & 90.9 & 88.2 & \textbf{99.9}  \\
		Tile & \textbf{99.5} & 61.3 & 79.1 & 59.0 & 97.3 & 65.0 & 97.8 & 99.1 & 98.7  \\
		Wood & 99.5 & 75.0 & 97.1 & 73.0 & \textbf{99.8} & 84.0 & 96.5 & 97.7 & 93.0  \\
		Bottle & \textbf{99.9} & 91.4 & 86.2 & 93.0 & 97.0 & 92.0 & 98.6 & 99.0 & \textbf{99.9}  \\
		Cable & 90.9 & 76.4 & 56.7 & 82.0 & 85.1 & \textbf{91.0} & 90.3 & 86.2 & 81.9  \\
		Capsule & 86.7 & 72.3 & 75.7 & \textbf{94.0} & 78.8 & 92.0 & 76.7 & 86.1 & 88.4  \\
		Hazelnut & \textbf{99.1} & 63.2 & 97.4 & 97.0 & 98.5 & 98.0 & 92.0 & 93.1 & 83.3  \\
		Metal nut & \textbf{97.0} & 59.7 & 53.1 & 89.0 & 76.1 & 91.0 & 94.0 & 82.0 & 88.5  \\
		Pill & 86.6 & 64.1 & 77.9 & 91.0 & 83.3 & \textbf{93.0} & 86.1 & 87.9 & 83.8  \\
		Screw & 89.6 & 50.0 & 83.6 & \textbf{96.0} & 82.4 & 95.0 & 81.3 & 54.9 & 84.5  \\
		Toothbrush & \textbf{99.9} & 67.3 & 96.8 & 82.0 & 96.9 & 98.0 & \textbf{99.9} & 95.3 & \textbf{99.9}  \\
		Transistor & \textbf{96.8} & 77.9 & 71.6 & 90.0 & 87.5 & 92.0 & 91.5 & 81.8 & 90.9  \\
		Zipper & \textbf{99.1} & 50.0 & 83.7 & 88.0 & 87.5 & 87.0 & 97.9 & 91.9 & 98.1  \\
		\midrule
		Textures & \textbf{98.32} & 63.00 & 89.04 & 78.20 & 97.74 & 82.40 & 94.54 & 91.52 & 95.08  \\
		Objects & \textbf{94.56} & 67.23 & 78.27 & 90.20 & 87.31 & 92.90 & 90.83 & 85.82 & 89.92  \\
		Mean & \textbf{95.81} & 65.82 & 81.86 & 86.20 & 90.79 & 89.40 & 92.07 & 87.72 & 91.64  \\
		\toprule
	\end{tabular}}
	\caption{{\bf{Detection results on MVTec AD (AUROC)}}. Ours outperforms other prior SOTA methods}\label{tabA5}}
\end{table*}
\begin{table*}\small
	\parbox{0.95\textwidth}{
		\resizebox{\linewidth}{!}{
	\begin{tabular}{ccccccccc}
		\toprule%
		Category&{\bf{Ours}} & AnoGAN & AESSIM & MEMAE & VAEGRAD &PatchSVDD &US &RIAD\\
		\midrule
		Carpet & 93.6  & 54.0  & 54.5  & 81.2  & 72.7  & 92.6  & 93.5  & \textbf{96.3}   \\
		Grid & 93.6  & 58.0  & 96.0  & 95.6  & 97.9  & 96.2  & 89.9  & \textbf{98.8}   \\
		Leather & 97.5  & 95.0  & 71.0  & 92.9  & 89.7  & 97.4  & 97.8  & \textbf{99.4}   \\
		Tile & \textbf{97.6}  & 93.0  & 49.6  & 70.8  & 58.1  & 91.4  & 92.5  & 89.1   \\
		Wood & \textbf{95.8}  & 91.0  & 64.1  & 85.4  & 80.9  & 90.8  & 92.1  & 85.8   \\
		Bottle & 98.2  & 86.0  & 93.3  & 85.0  & 93.1  & 98.1  & 97.8  & \textbf{98.4}   \\
		Cable & 95.8  & 78.0  & 79.0  & 71.2  & 88.0  & \textbf{96.8}  & 91.9  & 84.2   \\
		Capsule & 95.0  & 84.0  & 76.9  & 93.0  & 91.7  & 95.8  & \textbf{96.8}  & 92.8   \\
		Hazelnut & 98.3  & 87.0  & 96.6  & 97.2  & \textbf{98.8}  & 97.5  & 98.2  & 96.1   \\
		Metal nut & \textbf{98.0}  & 76.0  & 88.1  & 79.0  & 91.4  & \textbf{98.0}  & 97.2  & 92.5   \\
		Pill & 96.2  & 87.0  & 89.5  & 93.3  & 93.5  & 95.1  & \textbf{96.5}  & 95.7   \\
		Screw & 98.3  & 80.0  & 98.3  & 95.6  & 97.2  & 95.7  & 97.4  & \textbf{98.8}   \\
		Toothbrush & 98.4  & 90.0  & 97.3  & 94.8  & 98.3  & 98.1  & 97.9  & \textbf{98.9}   \\
		Transistor & \textbf{98.0}  & 80.0  & 90.4  & 66.7  & 93.1  & 97.0  & 73.7  & 87.7   \\
		Zipper & 97.1  & 78.0  & 82.8  & 84.6  & 87.1  & 95.1  & 95.6  & \textbf{97.8}   \\ 
		\midrule
		Textures & \textbf{95.62}  & 78.20  & 67.04  & 85.18  & 79.86  & 93.68  & 93.16  & 93.88   \\
		Objects & \textbf{97.33}  & 82.60  & 89.22  & 86.04  & 93.22  & 96.72  & 94.30  & 94.29   \\
		Mean & \textbf{96.76}  & 81.13  & 81.83  & 85.75  & 88.76  & 95.70  & 93.92  & 94.15   \\
		\toprule
	\end{tabular}}
	\caption{{\bf{Segmentation results on MVTec AD (AUROC)}}}\label{tabA6}}
\end{table*}
Some inspection results of our proposed methods on MVTec AD are shown in Figure~\ref{fig8}. These results show that the model can accurately inspect various types of defects, both textures and object defects.

\subsection{Applications}\label{subsec2}
\begin{figure*}[h]
	\centering
	\includegraphics[width=0.8\textwidth]{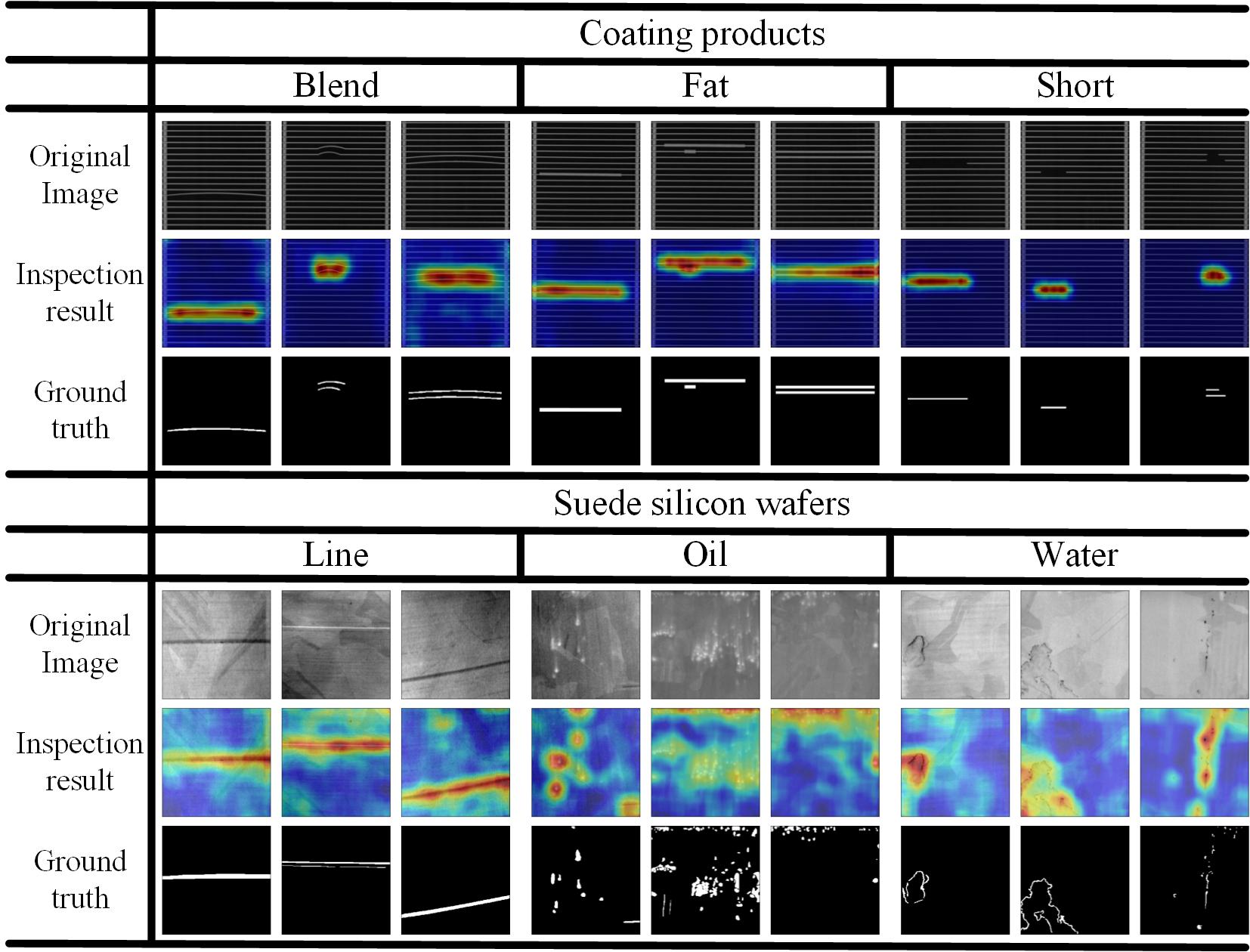}
	\caption{Anomaly maps of our real dataset by using our method}\label{fig10}
\end{figure*}

To further assess the practical performance of our method, we apply it to a dataset composed of texture and object samples in real industrial environments. This dataset contains two real-world products, coating products (CP) and suede silicon wafers (SSW) and there are about 200 training images and 100 testing images. The texture features of the SSW are complex and irregular, and there are significant similarities between the defects and textures.
The anomaly maps are described in Figure~\ref{fig10} in Sec.~\ref{sec:sup}, demonstrating the potential of our method in industrial applications. Moreover, the results of the comparison of different methods (AE-SSIM \cite{bib25}, US \cite{bib30}, RIAD \cite{bib31},PADIM \cite{bib32}, and SPADE \cite{bib33}) on our own dataset is shown in Table~\ref{table_real_data}.
\begin{table}
	\setlength\tabcolsep{5pt}
	\centering
	\resizebox{\linewidth}{!}{
		\begin{tabular}{c|cccccc}
			\specialrule{0.1em}{1pt}{1pt}
			& \footnotesize AESSIM	& \footnotesize RIAD	& \footnotesize US
			& \footnotesize PADIM	& \footnotesize SPADE & \footnotesize OURS\\
			\hline
			\scriptsize {Detection/CP}
			& \footnotesize97.8 & \footnotesize98.2 & \footnotesize98.4
			& \footnotesize99.7 & \footnotesize99.8 & \footnotesize \bf{99.9}\\
			\scriptsize {Segmentation/CP}
			& \footnotesize94.3 & \footnotesize95.5 & \footnotesize94.7
			& \footnotesize94.6 & \footnotesize94.8 & \footnotesize \bf{97.0}\\
			\hline
			\scriptsize {Detection/SSW}
			& \footnotesize67.8 & \footnotesize77.9 & \footnotesize83.6
			& \footnotesize99.3 & \footnotesize99.0 & \footnotesize \bf{99.9}\\
			\scriptsize {Segmentation/SSW}
			& \footnotesize64.2 & \footnotesize75.4 & \footnotesize82.3
			& \footnotesize87.5 & \footnotesize86.9 & \footnotesize \bf{89.0}\\
			\specialrule{0.1em}{1pt}{1pt}
		\end{tabular}
	}
	\vspace{-0.8em}
	\caption{{\bf{Different methods on real data.}} The AUCROC metric of detection and segmentation.}
	\label{table_real_data}
\end{table}

\section{Conclusion}
In this article, we propose a method for inspecting both texture and object defects in industrial applications, that is driven by a self-supervised learning. This method is trained on only a few defect-free samples. It utilizes context prediction to obtain a semantic encoder for better feature representations, which are then mapped into hyperspheres by using classic SVDD and stored in a memory module. Subsequently, defect detection and segmentation are conducted in the latent space by comparing the test image feature hyperspheres with the normal hyperspheres. To improve the feature matching accuracy, we propose an affinity-augmentation mechanism that obtains an aggregated contrastive pattern by weighting similar features. Extensive experimental results show that the proposed method achieves state-of-the-art inspection performance. In addition, image enhancement is suggested for extremely-slender or defects with extra-low contrast. In the future, we will apply our method as a defect inspection tool for more industrial applications.

{\small
	\bibliographystyle{ieee_fullname}
	\bibliography{reference}
}

\clearpage
\section{Supplementary Material}\label{sec:sup}
In the supplementary section, we will list the detail information mentioned above.

\begin{table}[h]
	\centering
	\begin{tabular}{cccc}
		\toprule%
		Layer & Output Size & Kernel & Stride \\
		\midrule
		Input  & $32\times32\times3$ & - & - \\
		Conv1  & $28\times28\times96$ & $5\times5$ & 1 \\
		Maxpool1  & $14\times14\times96$ & $3\times3$ & 2 \\
		Conv2  & $14\times14\times256$ & $5\times5$ & 1 \\
		Maxpool2  & $6\times6\times256$ & $3\times3$ & 2 \\
		Conv3  & $6\times6\times384$ & $3\times3$ & 1 \\
		Conv4  & $6\times6\times384$ & $3\times3$ & 1 \\
		Conv5  & $6\times6\times256$ & $3\times3$ & 1 \\
		Maxpool3  & $3\times3\times256$ & $3\times3$ & 2 \\
		Conv6  & $2\times2\times128$ & $2\times2$ & 1 \\
		Conv7  & $1\times1\times64$ & $2\times2$ & 1 \\
		\toprule
	\end{tabular}
\caption{Structure of the main encoder}\label{tabA1}
\end{table}

\begin{table}[h]
	\centering
	\begin{tabular}{cccc}
		\toprule%
		Layer & Output Size & Kernel & Stride \\
		\midrule
		Input  & $2\times2\times64$ & - & - \\
		Conv1  & $1\times1\times128$ & $2\times2$ & 1 \\
		Conv2  & $1\times1\times64$ & $1\times1$ & 2 \\
		\toprule
	\end{tabular}
	\caption{Structure of the secondary encoder}\label{tabA2}
\end{table}

\begin{table}[h]
	\centering
	\resizebox{\linewidth}{!}{
	\begin{tabular}{ccccccc}
		\toprule%
		{Category} & {Jigsaw1} & {Jigsaw2} & {Rotation} & {RGB} & {RR} & {Ours} \\
		\midrule
		Carpet & 92.9  & 82.6  & 61.1  & 77.7  & 85.6  & \textbf{95.7}   \\
		Grid & 94.6  & 93.1  & \textbf{99.4}  & 96.4  & 91.1  & 96.6   \\
		Leather & 90.9  & 88.6  & 96.6  & 94.8  & 92.9  & \textbf{98.8}   \\
		Tile & 97.8  & 98.9  & 96.1  & 98.9  & 98.2  & \textbf{99.3}   \\
		Wood & 96.5  & 98.0  & 94.2  & 92.7  & 94.7  & \textbf{99.1}   \\
		Bottle & 98.6  & 92.9  & 95.7  & 96.0  & 96.1  & \textbf{99.9}   \\
		Cable & 90.3  & 73.4  & \textbf{90.8}  & 73.6  & 86.2  & 90.7   \\
		Capsule & 76.7  & 79.9  & 85.7  & 80.8  & 77.6  & \textbf{86.4}   \\
		Hazelnut & 92.0  & 94.8  & 96.9  & 92.3  & 92.9  & \textbf{97.4}  \\
		Metal nut & 94.0  & 82.7  & 76.5  & 78.4  & 80.0  & \textbf{96.5}   \\
		Pill & 86.1  & 84.5  & 84.8  & \textbf{87.8}  & 86.3  & 86.4   \\
		Screw & 81.3  & \textbf{92.4}  & 85.9  & 80.0  & 85.8  & 89.2   \\
		Toothbrush & \textbf{100.0}  & 95.8  & 98.0  & 98.6  & 97.8  & 99.9   \\
		Transistor & 91.5  & 73.6  & 93.8  & 85.9  & 90.7  & \textbf{94.6}   \\
		Zipper & 97.9  & 89.3  & 92.1  & 90.0  & 88.3  & \textbf{98.8}   \\
		\midrule
		Textures & 94.54  & 92.24  & 89.48  & 92.10  & 92.50  & \textbf{97.90}   \\
		Objects & 90.84  & 85.93  & 90.02  & 86.34  & 88.17  & \textbf{93.98}   \\
		Mean & 92.07  & 88.03  & 89.84  & 88.26  & 89.61  & \textbf{95.29}   \\
		\toprule
	\end{tabular}}
	\caption{Comparison of the {\bf{detection}} results on MVTec AD (AUROC) of different self-supervised learning methods. Textures/Objects/Mean means the average AUROC of texture/object/all categories.}\label{tabA3}
\end{table}

\begin{table}[t]
	\centering
	\resizebox{\linewidth}{!}{
	\begin{tabular}{ccccccc}
		\toprule%
		{Category} & {Jigsaw1} & {Jigsaw2} & {Rotation} & {RGB} & {RR} & {Ours} \\
		\midrule
		Carpet & 92.6  & 93.8  & 75.4  & 94.5  & \textbf{97.4}  & 93.5   \\
		Grid & \textbf{96.2}  & 91.2  & 94.4  & 91.0  & 85.6  & 93.4   \\
		Leather & 97.4  & 96.0  & 97.4  & 95.5  & 95.4  & \textbf{97.5}   \\
		Tile & 91.4  & 95.4  & 95.0  & \textbf{98.1}  & 96.1  & 97.6   \\
		Wood & 90.8  & 92.7  & 91.0  & 89.5  & 90.8  & \textbf{95.8}   \\
		Bottle & 98.1  & 90.8  & 89.1  & 90.4  & 92.6  & \textbf{98.2}   \\
		Cable & \textbf{96.8}  & 91.0  & 92.6  & 91.2  & 92.7  & 95.8   \\
		Capsule & 95.8  & 96.8  & \textbf{96.9}  & 94.9  & 95.7  & 95.0   \\
		Hazelnut & 97.5  & 97.1  & 97.0  & 94.8  & 92.8  & \textbf{98.3}   \\
		Metal nut & \textbf{98.0}  & 93.4  & 70.5  & 95.4  & 90.7  & 97.8   \\
		Pill & 95.1  & 96.0  & 96.5  & \textbf{97.6}  & 96.2  & 96.2   \\
		Screw & 95.7  & 96.4  & 96.3  & 93.8  & 96.3  & \textbf{98.3}   \\
		Toothbrush & 98.1  & 97.2  & 97.7  & 97.6  & \textbf{98.4}  & 98.2   \\
		Transistor & \textbf{97.0}  & 94.7  & 96.1  & 92.0  & 90.4  & 96.6   \\
		Zipper & 95.1  & 94.9  & 94.9  & 94.6  & 92.3  & \textbf{97.0}   \\
		\midrule
		Textures & 93.68  & 93.82  & 90.64  & 93.72  & 93.06  & \textbf{95.56}   \\
		Objects & 96.72  & 94.83  & 92.76  & 94.23  & 93.81  & \textbf{97.14}   \\
		Mean & 95.71  & 94.49  & 92.05  & 94.06  & 93.56  & \textbf{96.61}   \\
		\toprule
	\end{tabular}}
	\caption{Comparison of the {\bf{segmentation}} results on MVTec AD (AUROC) of different self-supervised learning methods}\label{tabA4}
\end{table}

\end{document}